\documentclass[runningheads]{llncs}
\usepackage{graphicx}

\usepackage{tikz}
\usepackage{comment}
\usepackage{amsmath,amssymb} 
\usepackage{color}

\usepackage{booktabs}
\usepackage{soul}
\usepackage{multirow}
\usepackage{multicol}

\usepackage[accsupp]{axessibility}  
\usepackage{orcidlink}

\newcommand{\bez}{Bézier}

\newcommand{\tr}{toprock}
\newcommand{\fw}{footwork}
\newcommand{\pwm}{powermove}
\newcommand{\fparagraph}[1]{\noindent\textit{#1}}
\newcommand{\etal}{\textit{et al.}}

\begin{document}
\pagestyle{headings}
\mainmatter
\def\ECCVSubNumber{5564}  

\title{BRACE: The Breakdancing Competition Dataset for Dance Motion Synthesis} 

\titlerunning{BRACE: The Breakdancing Competition Dataset}
%
\author{Davide Moltisanti\inst{1}\thanks{Work done while at Nanyang Technological University. \textsuperscript{$\dagger$} Equal contribution.}\textsuperscript{$\dagger$}\orcidlink{0000-1111-2222-3333}\index{Moltisanti, Davide} \and
Jinyi Wu\inst{2}\textsuperscript{$\dagger$}\orcidlink{0000-0002-1869-9396}\index{Wu, Jinyi} \and
Bo Dai\inst{3}\textsuperscript{$\star$}\orcidlink{0000-0003-0777-9232}\index{Dai, Bo} \and
Chen Change Loy\inst{4}\orcidlink{0000-0001-5345-1591}\index{Loy, Chen Change}}
\authorrunning{D. Moltisanti et al.}
%
\institute{University of Edinburgh, \email{davide.moltisanti@ed.ac.uk} \and
S-Lab, Nanyang Technological University, \email{jinyi002@e.ntu.edu.sg} \and
Shanghai AI Laboratory, \email{daibo@pjlab.org.cn} \and
S-Lab, Nanyang Technological University, \email{ccloy@ntu.edu.sg}}

\maketitle

\begin{figure*}[t]
    \centering
    \includegraphics[width=\linewidth]{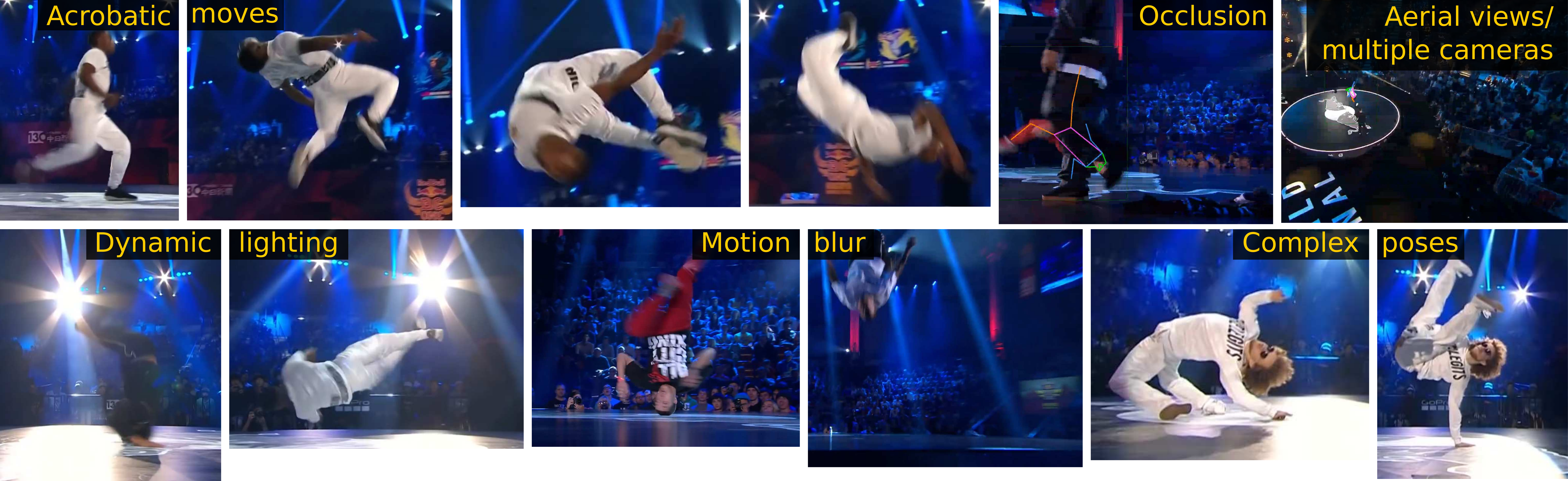}
    \caption{Challenges and characteristics of BRACE's video source. Acrobatic moves make our dataset unique compared to previous dance datasets. Complex poses are difficult to estimate automatically. Occlusion, multiple cameras setup with frequent shot changes and aerial views, dynamic lighting with strong flashes and motion blur add to the difficulty of extracting keypoints for the dancer.}
    \label{fig:challenges}
\end{figure*}

\setcounter{footnote}{0} 

\begin{abstract}
Generative models for audio-conditioned dance motion synthesis map music features to dance movements. 
Models are trained to associate motion patterns to audio patterns, usually without an explicit knowledge of the human body. This approach relies on a few assumptions: strong music-dance correlation, controlled motion data and relatively simple poses and movements. 
These characteristics are found in all existing datasets for dance motion synthesis, and indeed recent methods can achieve good results.
We introduce a new dataset aiming to challenge these common assumptions, compiling a set of dynamic dance sequences displaying complex human poses. We focus on breakdancing which features acrobatic moves and tangled postures. We source our data from the 
Red Bull BC One competition videos. Estimating human keypoints from these videos is difficult due to the complexity of the dance, as well as the multiple moving cameras recording setup. We adopt a hybrid labelling pipeline leveraging deep estimation models as well as manual annotations to obtain good quality keypoint sequences at a reduced cost. 
Our efforts produced the BRACE dataset, which contains over 3 hours and 30 minutes of densely annotated poses. We test state-of-the-art methods on BRACE, showing their limitations 
when evaluated on complex sequences. 
Our dataset can readily foster advance in dance motion synthesis. 
With intricate poses and swift movements, models are forced to go beyond learning a mapping between modalities and reason more effectively about body structure and movements.
\end{abstract}

\section{Introduction}
\label{sec:intro}

Audio-conditioned generative models~\cite{tang2018dance,zhuang2020music2dance,huang2021,li2021learn} for dance motion synthesis learn a mapping between music features  
and the aligned dance movements. Models typically tackle this task treating the skeleton sequence as an additional modality to be ``remembered'' when ``hearing'' a particular music beat or melody. Models are  
trained to capture correspondences between a sequence of music features and a sequence of 
keypoints, often without explicitly modelling the body's structure and movements. 
This is a reasonable approach when the following conditions are met: i) there is a strong correlation between music and dance movement, i.e. a choreography can be observed; ii) poses and movements are relatively simple or not too diverse and iii) motion data is controlled, i.e. all the captured movement is related to the dance, without keypoint shifts induced by camera movements. 
Current datasets~\cite{alemi2017groovenet,tang2018dance,lee2019dancing2music,zhuang2020music2dance,li2021learn,huang2021,li2021dancenet3d,ferreira2021learning} satisfy these constraints, where videos are captured either in a controlled environment~\cite{alemi2017groovenet,tang2018dance,zhuang2020music2dance,li2021learn}, with a static camera ~\cite{lee2019dancing2music,ferreira2021learning,huang2021} or are even synthesised~\cite{li2021dancenet3d}. 

We propose a new dataset aiming to challenge such assumptions. Our goal is to gather data that: i) contains diverse and complex human poses and movements; ii) exhibits a weaker correlation between music and movement; iii) is captured in a real-world setting, i.e. multiple moving cameras record dancers from a variety of viewpoints. 
We look after these characteristics to push generative models to go beyond learning a mapping between modalities. With weak music-dance correlation, less controlled motion and difficult movements and poses, models are forced to effectively reason about body structure and movement. 
The properties we are seeking are amply available in videos of breakdancing, which is an athletic form of dance with swift movements and tangled poses. Music is typically a looping sequence providing a rhythmic beat  
rather than a base for a choreography, and as such it is weakly correlated with the dance. 

We use videos from the Red Bull BC One contests featuring the best dancers in the world.  
Thanks to recent progress in deep pose estimation models, keypoints are typically entirely estimated, and good results can be achieved when  
poses are relatively simple. 
However, the dynamic nature of breakdancing and the recording setting in our video source introduce several challenges to keypoint estimation. Due to the complex posture assumed by the dancers, pose estimation models are pushed to their limits. Motion blur induced by fast and acrobatic moves, occlusion and dynamic lighting further complicate the task, as we show in Figure~\ref{fig:challenges}. Under these circumstances, it is not possible to solely rely on pose estimation models to obtain good quality keypoints. At the same time, manually annotating poses in videos is very expensive and time consuming. To overcome these issues we design a hybrid annotation pipeline using both automatic and manual annotations, striking a good balance between keypoint quality and~labelling~burden. 
Our efforts produced the BRACE dataset, a collection of dynamic dance sequences annotated with high quality 2D keypoints. As reported in Table~\ref{tab:dataset_info}, our dataset amounts to over 3 hours and 30 minutes of footage.
We also provide fine-grained labels annotating the key elements of breakdancing. We test recent state-of-the-art methods on BRACE, showing the limitations of current approaches when evaluated on more dynamic and complex data. 

To summarise, our contributions are: i) a new high quality dataset featuring complex poses and dynamic movements; ii) a hybrid automatic-manual annotation pipeline designed to efficiently annotate human keypoints under challenging conditions; iii) a study of recent work on our new dataset, showing the challenges posed by its unique characteristics.
BRACE is publicly available online \footnote{\url{https://github.com/dmoltisanti/brace/}}.

\section{Related Work}
\label{sec:related_work}

\subsection{Dance Datasets}
\label{sec:dataset_comparison}

\begin{table*}[t]
\centering
\caption{Comparing recent datasets for dance motion synthesis. A: automatic (estimated with a model). M: manually annotated. C: obtained from MoCap. *not available. 
}
\resizebox{\textwidth}{!}{%
\begin{tabular}{@{}lcccccccccc@{}}
\toprule
\textbf{Name} & \textbf{Year} & \textbf{Sequences} & \textbf{Size} & \textbf{Dancers} & \textbf{Styles} & \textbf{Annotations} & \textbf{Source} & \textbf{FPS} & \textbf{Movement} & \textbf{Pose} \\ \midrule
Groove net~\cite{alemi2017groovenet} & 2017 & 4 & 0h 23m & 1 & 1 & 3D joints (C) & Hired dancer & NA* & NA* & NA* \\ \midrule
Dance with melody~\cite{tang2018dance} & 2018 & 61 & 1h 34m & NA & 4 & \begin{tabular}[c]{@{}l@{}}3D joints (C),\\ basic dance movements\end{tabular} & Hired dancers & 25 & 1.918 & 0.193 \\ \midrule
Dancing 2 Music~\cite{lee2019dancing2music} & 2019 & 361K & 71h & NA & 3 & 2D keypoints (A) & YouTube & 15 & 0.704 & 0.071 \\ \midrule
Music 2 Dance~\cite{zhuang2020music2dance} & 2020 & 2 & 0h 58m & 2 & 2 & 3D joints (C) & Hired dancers & 60 & 1.583 & 0.205 \\ \midrule
AIST++~\cite{li2021learn} & 2021 & 1.4K & 5h 11m & 30 & 10 & \begin{tabular}[c]{@{}l@{}}2D/3D keypoints and\\ SMPL body models (A)\end{tabular} & Hired dancers & 60 & 1.450 & 0.135  \\ \midrule
Dance revolution~\cite{huang2021} & 2021 & 790 & 12h & NA & 3 & 2D keypoints (A) & YouTube & 15 & 0.804 & 0.137 \\ \midrule
Phantom Dance~\cite{li2021dancenet3d} & 2021 & 300 & 12h & 1 & 4 & 3D quaternion & Synthesised & 60 & 1.962 & 0.251 \\ \midrule
Learning to Dance~\cite{ferreira2021learning} & 2021 & 298 & 0h24m & NA & 3 & 2D keypoints (A) & YouTube & 24 & 1.161 & 0.111 \\ \midrule
\textbf{BRACE (Ours)} & 2022 & 465 & 3h 32m & 64 & 1 & \begin{tabular}[c]{@{}l@{}} 2D keypoints (A/M),\\movement categories,\\dancer IDs, shot bounds\end{tabular} & YouTube & 25-30 & 2.388 & 0.235 \\\bottomrule
\end{tabular}%
}
\label{tab:datasets}

\end{table*}

Table~\ref{tab:datasets} reports recent datasets for music conditioned dance motion generation. Earlier efforts~\cite{alemi2017groovenet,tang2018dance} collect a small amount of highly curated data, with motion capture setups recording hired dancers. 
AIST++~\cite{li2021learn} offers 2D/3D keypoints and SMPL models extracted for a subset of AIST~\cite{tsuchida2019aist}.  
Synthetic datasets have also been proposed~\cite{li2021dancenet3d}, where digital artists produce video animations of dance videos.
As pose estimation models grow increasingly reliable, sourcing videos from YouTube and extracting keypoints has become a common option~\cite{lee2019dancing2music,huang2021,ferreira2021learning}. 
We partially follow this approach, sourcing videos from YouTube and using a hybrid approach combining automatic and manual annotations.

A common characteristic of current datasets is their relative simplicity of pose and motion.  
Dancer in datasets such as Dancing 2 Music~\cite{lee2019dancing2music}, Dance Revolution~\cite{huang2021} and Learning to Dance~\cite{ferreira2021learning} are mostly in an upright position and do not perform particularly dynamic moves.  
AIST++~\cite{li2021learn} includes breakdancing, however these amount to only 30 minutes of footage (10\% of the whole dataset). Furthermore, based on our observations breakdancing sequences in AIST++ 
mostly feature simple moves and upright poses. We next quantify the differences amongst these datasets and BRACE measuring movement and pose diversity. 

\fparagraph{Measuring movement and pose diversity} 
We need to normalise all keypoints in a consistent way for a fair comparison.
We also wish to  marginalise apparent movements that could be induced by a camera change, zoom or panning.  
We extract the tightest box/cuboid enclosing all keypoints and normalise each keypoint with respect to the tightest box/cuboid. For example in 2D, given the tightest box $B = (x_b, y_b, w, h)$ and a keypoint $P = (x_p, y_p)$, the normalised keypoint is $\bar{P} = ((x_p - x_b)/w, (y_p - y_b)/h)$.  
To measure movements we take into account the FPS of the videos, since a small FPS may produce an unrealistic large displacement and vice-versa. For a given node $p$ its movement is thus calculated as $v_i = d(p_i, p_{i+1})/dt$, where $d(p_i, p_{i+1})$ is the Euclidean distance between frames $i$ and $i+1$ and $dt=1/fps$. We measure such frame-wise distance for each node independently and take the average on all sequences in a given dataset. This is what we report under ``Movement'' in Table~\ref{tab:datasets}. 
To measure pose diversity we calculate the standard deviation of each node. 
We calculate the $std$ of each node in a sequence and take the average across all sequences in a dataset and across all nodes. This gives us a 2D or 3D vector depending on the dimension of the keypoints. We report the average of this vector under ``Pose'' in Table~\ref{tab:datasets}.
To further reduce camera movement bias, we calculate metrics within shot boundaries.
As the table 
shows, our dataset offers by far the most dynamic sequences thanks to the very nature of breakdancing. While Phantom Dance exhibits a slightly larger pose diversity, it should be noted that this is a synthesised dataset.

\subsection{Dance Motion Synthesis}
\label{sec:rel_work_motion_syn}

\fparagraph{Sequence-to-sequence approaches}~\cite{tang2018dance,zhuang2020music2dance,huang2021,li2021learn} are a common choice for audio-conditioned dance motion generation. 
Huang \etal~\cite{huang2021} design an Encoder-Decoder (ED) architecture, where
the Encoder is a transformer that encodes music features in a latent space. The decoder is an RNN that produces the dance movement as a sequence of keypoints. 
Li \etal~\cite{li2021learn} also adopt a sequence-to-sequence framework, using a group of transformers.  
When motion data is controlled 
a 
sequence mapping can successfully produce good dance motion by implicitly learning movements from the data. 
However, when data is less controlled,  
i.e. data captures motion induced by camera movements (e.g. zooming, panning, camera change), we expect models to produce similar artefacts since they learn to replicate the training sequences. While normalisation techniques  
can mitigate this issue, ultimately it becomes harder for a model to learn motion when this is captured ``in the wild''. 
When poses and motion are more diverse and complex the mapping task also becomes intrinsically harder. We test~\cite{huang2021,li2021learn} on BRACE, where these challenges are widely present, to study the limitations of sequence-to-sequence approaches. 

\fparagraph{Other approaches}
~\cite{lee2019dancing2music} extracts music and dance units from the data. Music units are extracted with an audio beat detection algorithm. Dance units are found detecting abrupt changes in motion magnitude and angle. 
Dance sequences are decomposed into such units.
An auto-encoder models these atomic moves, and a GAN composes multiple units to form a dance sequence. 
The motion unit decomposition of this method is likely to be challenged when dance sequences exhibit swift movements and acrobatic poses as in BRACE, as we show in Section~\ref{sec:testing_sota}.
In 
~\cite{li2021dancenet3d} an ED generates key poses, while another ED generates motion curves between these key poses. 
The authors propose a Kinematic Chain Network where features are embedded with MLPs that are chained following the body structure, using a tree topology.  
In ~\cite{yan2019convolutional} a latent space is constructed sampling from a Gaussian process. The latent space encodes an abstract motion signal, and a GCN is then trained to map such signal to a skeleton sequence. 
\cite{yan2019convolutional} does not condition motion generation on music, which is done in~\cite{ferreira2021learning} through a GCN-based generative model trained in an adversarial fashion.  

\section{The BRACE Dataset}
\label{sec:data_description}

\begin{table}[t]
\centering
\caption{The BRACE dataset at a glance. 
Each dancer performs multiple sequences in a video. 
Sequences are annotated into segments according to their dance element.}
\resizebox{0.75\columnwidth}{!}{%
\begin{tabular}{@{}llll@{}} \toprule
\textbf{Frames} & 334,538 & \textbf{Sequences} & 465 \\
\textbf{Manually annotated frames} & 26,676 (8\%) & \textbf{Segments} & 1,352 \\
\textbf{Duration} & 3h 32m & \textbf{Avg. segments per sequence} & 2.91 \\
\textbf{Dancers} & 64 & \textbf{Avg. sequence duration} & 27.48s \\
\textbf{Videos} & 81 & \textbf{Avg. segment duration} & 9.45s \\ \bottomrule
\end{tabular}%
}
\label{tab:dataset_info}

\end{table}

We choose the Red Bull BC One breakdancing competition as our data source, 
which
features the best dancers in the world competing one against each other. The competition follows the knockout tournament format, where two dancers compete in a 1-vs-1 battle taking turns to perform a number of sequences. 
Video recordings of the shows are available on YouTube~\footnote{\url{https://www.youtube.com/channel/UC9oEzPGZiTE692KucAsTY1g}} (we use videos from years 2011/13/14/17/18/20).
Figure~\ref{fig:challenges} illustrates the videos characteristics and the difficulties involved in extracting human poses. Videos were shot using multiple cameras and feature wide panning, long zooming, aerial views and abrupt shot changes, all of which makes automatic pose extraction not trivial. Lighting is also a challenge, since strong flashes and dimly lit scenes are common in the videos. Furthermore, the very nature of breakdancing makes keypoint extraction intrinsically difficult. Complex poses and motion blur induced by quick movements stretch the capabilities of pose estimation models.
We design a pipeline to address such challenges to extract 2D keypoints from the videos, adopting a hybrid annotation paradigm. Our objective is to obtain dense and good quality poses while minimising manual annotations cost. Our pipeline is flexible and can be adopted for other pose estimation datasets. 

It should be noted that we are not affiliated with Red Bull and that all video-audio copyright belongs to Red Bull. We release our processed keypoint sequences and our temporal labels, providing a link to the original YouTube videos. We also release audio features extracted with Librosa~\cite{mcfee2015librosa}.

\subsection{Data Acquisition}
\label{sec:pipeline}

\begin{figure*}[t]
    \centering
    \includegraphics[width=\linewidth]{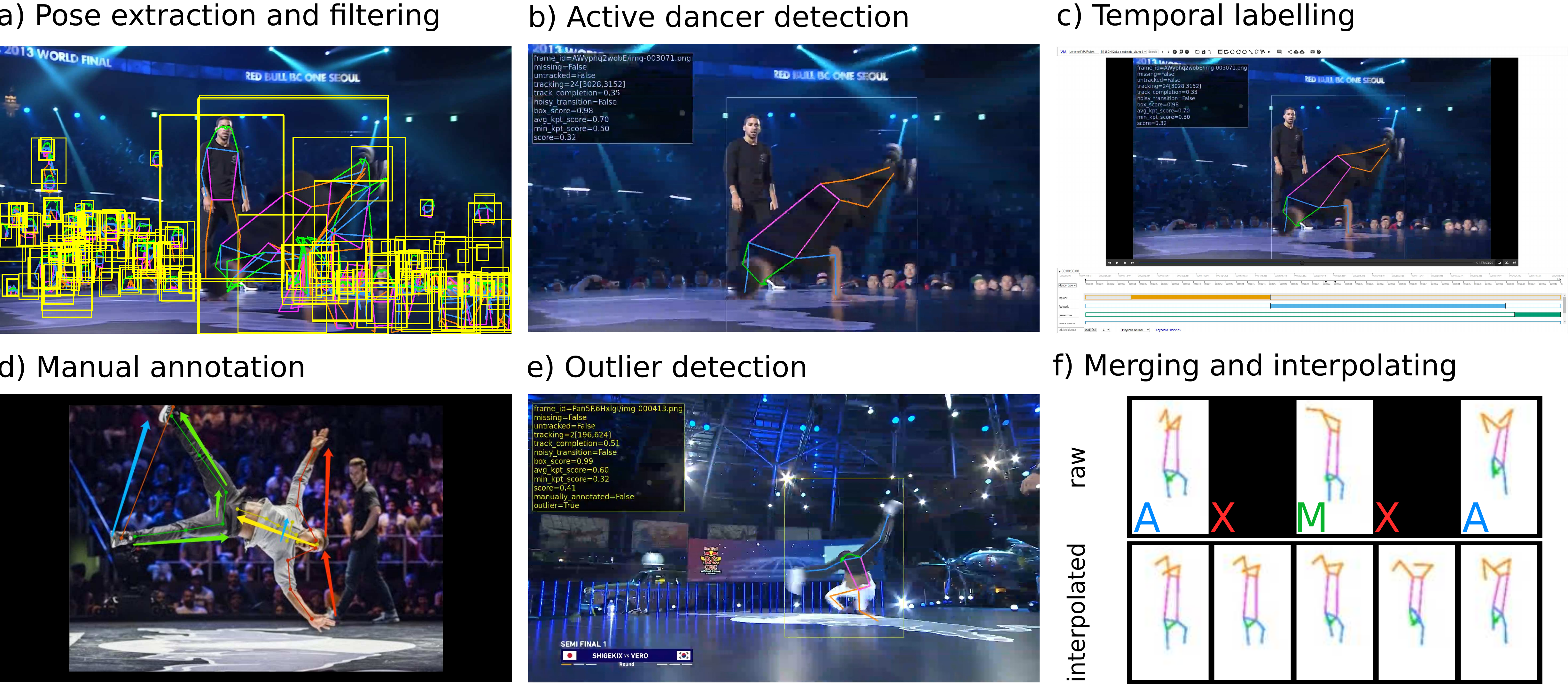}
    \caption{Data acquisition pipeline. We start extracting automatic poses with a model ensemble, which produces a large number of annotations (a). We then filter poses and detect the active dancer throughout the video (b) before annotating dance segments (c). Subsequently, we select bad poses and manually label them (d). Automatic pose outliers are then detected (e) and manually labelled. Finally, manual and automatic pose are merged and interpolated to produce our dance keypoint sequences (f, where ``A'', ``X'' and ``M'' indicate automatic, missing and manual poses respectively).}
    \label{fig:pipeline}
\end{figure*}

Figure~\ref{fig:pipeline} shows our approach to obtain human keypoints for the active dancer. We first employ state-of-the-art human pose estimators to extract automatic poses. We then process keypoints to select the active dancer and temporally annotate dance sequences.
As discussed earlier, 
we cannot solely rely on automatic pose estimation. We thus find bad keypoints corresponding to difficult poses and manually label them. We later detect automatic pose outliers and also manually label these. Finally, we merge automatic and manual annotations interpolating the keypoint sequences with \bez{} curves. Table~\ref{tab:dataset_info} summarises our dataset. 

\fparagraph{Automatic pose estimation} We employ a model ensemble to extract human poses from video frames. We favour this option over a single model to boost the chance of getting good keypoints for difficult poses. We choose a top-down approach to first detect humans in frames and then estimate their pose, using MMDetection~\cite{MMDetection_Contributors_OpenMMLab_Detection_Toolbox_2018} and MMPose~\cite{MMPose_Contributors_OpenMMLab_Pose_Estimation_2020}. We use the top 3 human detectors 
ranked on the COCO dataset~\cite{lin2014microsoft}: HTC~\cite{chen2019hybrid} with backbones ResNet 50/100~\cite{he2016deep} and Cascade R-CNN~\cite{Cai_2019} with backbone ResNet 50. Accordingly, we utilise the top 3 pose estimators 
ranked on COCO: HRNet~\cite{sun2019deep} and a cascaded HRNet+DarkPose~\cite{Zhang_2020_CVPR} with two different configurations (32 and 64 channels for the bottleneck block). All models were pre-trained on COCO. We obtain poses for each detector/estimator combination, gathering a total of 9 sets of poses for each frame. 

\fparagraph{Filtering multiple pose estimations} 
We start rejecting poses 
with low confidence scores.
We keep the 4 largest boxes produced by each model at a given frame, which removes most poses captured for people in the audience. Finally we apply a standard object-detection NMS. 
Figure~\ref{fig:pipeline} (a) shows all poses detected with the model ensemble before filtering. Notice the abundant number of detections. Figure~\ref{fig:pipeline} (b) illustrates the automatically selected pose for the active dancer.

\fparagraph{Active dancer detection}  
We automatically find the active dancer by building people tracks. These are obtained by linking boxes across frames if their IOU is above a certain threshold (0.4). For robustness we look for overlapping boxes in a window of 10 frames, i.e. given a box at frame $t$, if there is no overlapping box at frame $t+1$ we search for a box in frames $[t+2, t+10]$, choosing the temporally closest overlapping box. This simple method based entirely on boxes is good enough for our purpose. 
We measure the movement of each tracked person by calculating the change in pixel area of the tracked box across frames. 
We select the box corresponding to the track exhibiting the highest movement as our active dancer, and use the corresponding keypoints as the pose candidate. 

\fparagraph{Temporal labelling} We generate a new copy of each video overlaying the keypoints of the detected active dancer. We then label the start/end of each dance element, annotating the type of movement (toprock, footwork and powermove, presented in Section~\ref{sec:dance_elements}) as well as the dancer ID. We also annotate segments where the active dancer estimation was incorrect, i.e. frames where the pose candidate corresponded to the idle dancer or another person in the video. We use these labels to later retrieve keypoints for the correct person automatically. 

\fparagraph{Manual pose annotation} The quality of keypoint estimation degrades when the dancer's pose is extreme or when frames are too blurry. 
We manually label such difficult poses. To automatically find which frames need manual annotation we aggregate a labelling score. This is obtained multiplying the candidate's box confidence score and the average keypoint confidence score (a low labelling score indicates the pose is likely bad). We treat frames with a low labelling score as missing frames we need to manually annotate. 
We then apply a labelling discount to reduce annotation time and cost. Thanks to our Bézier interpolation, we can allow for a certain number of missing frames in a given window, while still recovering a good keypoint sequence. 
We allow for a maximum of 2 consecutive and a total of 5 missing frames in a window of 10 frames. The selected frames are then labelled by our locally sourced annotators. With our discount method we annotated only 57.17\% of frames initially marked for labelling. 

\fparagraph{Outlier detection} Some bad automatic poses may not be selected for manual annotations due to a noisy high confidence score.  
We find such poses with an outlier detection method. Once we have manually annotated frames for a video, we merge automatic and manual poses. We then employ a sliding window median filter to detect noisy outliers.  
Figure~\ref{fig:pipeline} (e) shows an outlier found with this algorithm. Green segments indicate the head, thus the person is incorrectly detected in an upright position. We search for outliers within the labelled segments and shot boundaries (detected with~\cite{scenedetect}). All outliers are manually labelled. 

\fparagraph{Merging and interpolating poses} At this point we have a mix of manual and automatic annotations, with missing poses for a few frames due to our labelling discount. In order to obtain smooth dance sequences we interpolate the keypoint series using \bez{} curves. We slide an overlapping window throughout a sequence of keypoints and fit a \bez{} curve to each node trajectory separately.  
Curves are fitted using a least-square algorithm. Sequences are interpolated within dance segments and shot boundaries. The interpolated sequences constitute our final keypoints data. More details can be found in the supplementary material.

\begin{table}[t]
\centering
\caption{Quality control measures for BRACE. Incorrect pose shows the percentage of the whole dataset where we spotted wrong keypoints. GT/Raw - Interpolated MAE refers to the Mean Absolute Error calculated for a fully manually annotated sequence.}
\label{tab:quality_check}
\resizebox{0.6\textwidth}{!}{%
\begin{tabular}{@{}ccccc@{}}
\toprule
\multicolumn{2}{c}{\textbf{Incorrect pose}} & \multicolumn{2}{c}{\textbf{GT - Interpolated MAE}} & \textbf{GT - Raw MAE} \\ \midrule
Automatic & 
Manual  & 
All frames & 
Rejected frames & 
Rejected frames\\
0.63\% & 0.12\% & 27px & 35px & 60px \\ \bottomrule
\end{tabular}%
}
\end{table}

\subsection{Quality Control} We carefully reviewed all the final interpolated sequences 
to spot incorrect poses. This is to validate the automatic selection of bad poses as well as the manual annotations themselves. 
We generated and watched videos overlaying the keypoints, marking all frames with a bad pose.
An odd-looking frame in the interpolated sequence is introduced when automatic or manual keypoints are incorrect. Table~\ref{tab:quality_check} reports how many of these we spotted. In total 0.63\% and 0.12\% of the total frames were found incorrect for automatic and  manual keypoints. 
These very low error percentages reflect the reliability of our data acquisition pipeline. Wrong automatic keypoints were outliers that were not detected. This typically happens when many consecutive poses are noisy, i.e. a bad pose is no longer an outlier in a window of frames according to the rolling window median filter. This can be fixed by tuning the outlier detector parameters. Manual incorrect annotations were found over frames with severe occlusion or where annotators labelled the idle dancer. This issue can be alleviated by showing annotators a few neighbouring frames in addition to the frame to be labelled.

We also implemented another quality control measure. We manually labelled a sequence of
1,472 frames (1min). The sequence contains toprocks, footworks and powermoves. The Mean Absolute Error (MAE, averaged across frames and joints) between the GT and the interpolated sequence is 27px (frame area: 1920x1080), as reported in Table~\ref{tab:quality_check}. The low MAE is indicative of BRACE's quality. 355 frames (24\%) of the raw keypoints were rejected (bad pose). 7\% of these frames were manually annotated for the interpolated sequence. Looking only at the rejected frames (i.e. those with a difficult pose), the GT-raw MAE is 60px whereas the GT-interpolated MAE is 35px. This further highlights the efficiency of our pipeline to obtain good poses labelling  
fewer frames ({7 vs 24\%}).

\subsection{Dance Elements}
\label{sec:dance_elements}

\begin{figure*}[t]
    \centering
    \includegraphics[width=\linewidth]{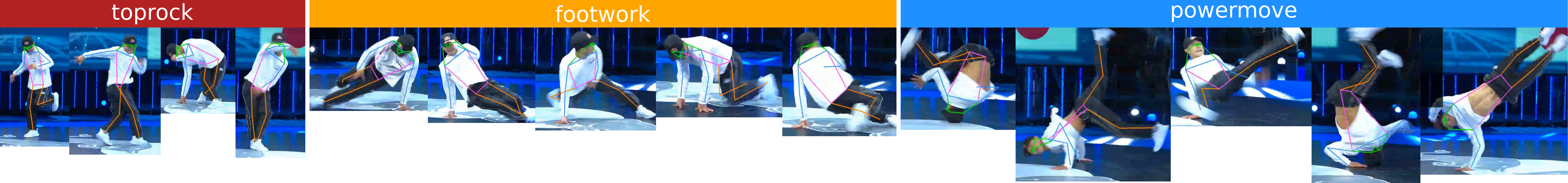}
    \caption{The primary elements of breakdancing: toprock, footwork and powermove.}
    \label{fig:movements}
\end{figure*}

A breakdancing sequence can be decomposed into three main elements: toprock, footwork and powermove. We label sequences into such constituent parts, and here we describe them briefly. A toprock features a dancer in the upright position performing a variety of free-style steps. A footwork involves a dancer supporting themselves on the floor using their hands while performing moves with their legs and feet. Powermoves are the most dynamic movements where performers engage in complex moves such as acrobatic aerial flips and head/back spins. Figure~\ref{fig:movements} illustrates a few frames for each dance element. Note the complexity of the dancer poses, especially for powermove and footwork. Next, we analyse how such movements are combined temporally.

 \begin{figure*}[t]
     \centering
     \includegraphics[width=\linewidth]{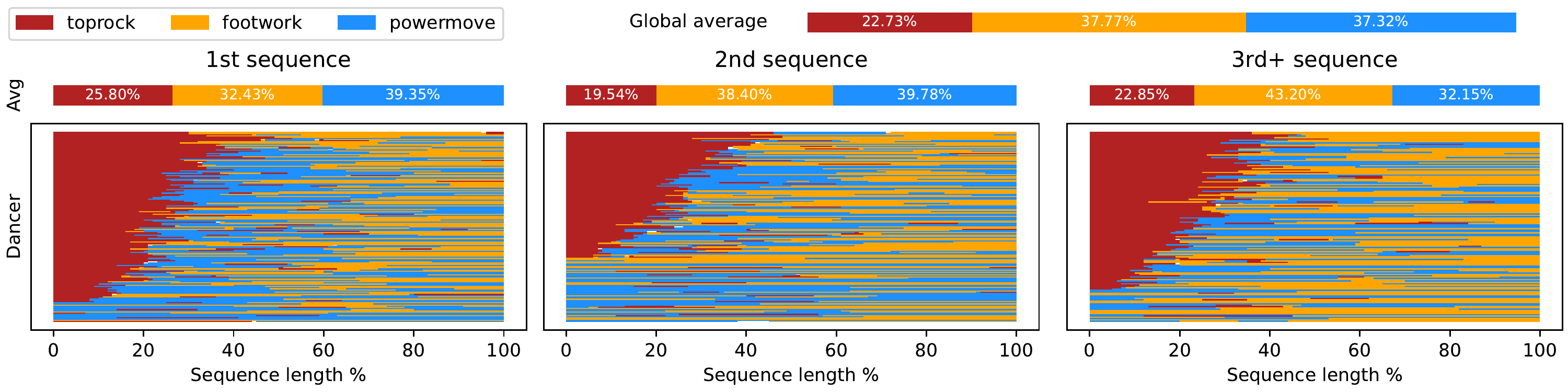}
     \caption{Analysing dance elements patterns emerging from the battle-format. We split sequences according to their order in the battle. Bottom: each line shows a sequence from a single dancer, with coloured segments corresponding to the three different movements we labelled. Segments are normalised according to the sequence length. We report the average percentage of frames labelled with each movement, both according to the sequence order (top) and globally (top right).}
     \label{fig:sequence_inspection}
 \end{figure*}

\fparagraph{Cypher format} 
We investigate here whether the alternating battle format brings any patterns in the dance. 
In Figure~\ref{fig:sequence_inspection} we plot each sequence according to their order in the competition. For each sequence we illustrate the segments annotated with the corresponding dance element (bottom of the figure). We observe that the vast majority of dancers begins their competition (1st sequence) with \tr{} movements, which is a common practice in breakdancing. Some dancers though do not follow such convention and start off with \pwm{}s. We also notice that the most common pattern in a sequence is (\tr, \pwm, \fw), which is interestingly more evident in the 1st sequence. Figure~\ref{fig:sequence_inspection} (top) also illustrates how often each movement was performed in a sequence (calculated as average percentage of the total length). We again see that the highest concentration of \tr{} appears in the 1st sequence. 
The percentage of \fw{} grows according to the order of the sequence. Interestingly, \pwm{} percentage is virtually unchanged in the first two sequences but drops by 7\% in the last sequence, which suggests that dancers become more tired towards the end of the competition, given that \pwm{}s are very strenuous. Finally, Figure~\ref{fig:sequence_inspection} (top right) reports global movement percentages.  
Footwork and \pwm{} are the dominant movements, confirming the dynamism of BRACE. We will look at global movement percentages again when testing our dataset. 

We believe the alternating battle format is an interesting novel aspect of BRACE. Indeed, by considering a sequence not just as a stand-alone progression of movements but as part of a longer and more complex scenario one could devise more creative models simulating dance competitions. While here we focus on breakdancing competitions, other styles too are often performed following such battle format (e.g. modern styles such as Hip-Hop, Krump and Street Jazz). 

\subsection{Audio Correlation}
\label{sec:audio_corr}

Although breakdancing has a weaker audio correlation, 
dancers still closely follow the music and its rhythm. Quoting from Red Bull BC One website \footnote{\label{redbullwebpage}\url{https://www.redbull.com/us-en/understand-the-basic-elements-of-breaking}}, dancers use especially toprock movements to ``\textit{showcase their rhythmic style and their ability to play with the music}''. As another example, dancers also follow the music to perform a freeze. Quoting again from \textsuperscript{\ref{redbullwebpage}}: ``\textit{a freeze is when a dancer holds a solid shape with their body for a few seconds. This is usually done to hit a prominent sound in the music}''. 
While the \textit{melody} of the music might be less correlated with the dance, the \textit{rhythm} is the foundation of the movements. Models are pushed to their limits,  
however 
the fundamental premise of the audio-conditioned generative task remains solid, i.e. there is sufficient correlation in the data for a model to learn to generate sequences based on the music.

Finally, we note that audio files contain live commentary of the performances. Theoretically, these could provide cues models could exploit. We watched 8 random full videos from all years (23 minutes of footage, 11\% of the dataset) and labelled segments with commentary. We found that 17\% of the labelled sequences contained a vocal comment of the performance. 
Based on these numbers, we conclude commentary is present but not likely to influence generative models.

\section{Testing Generative Models on BRACE}
\label{sec:testing_sota}

Testing models that were designed with specific assumptions on data where such constraints are more loose might not appear fair. Our intention is to test a new benchmark to encourage novel approaches to tackle more challenging data, rather than cast a bad light onto existing work. Here we evaluate Dance revolution~\cite{huang2021}, AIST++~\cite{li2021learn} and Dancing 2 Music~\cite{lee2019dancing2music}, reviewed in Section~\ref{sec:rel_work_motion_syn}.

\subsection{Evaluation Metrics} 

\begin{figure}[t]
    \centering
    \includegraphics[width=0.5\columnwidth]{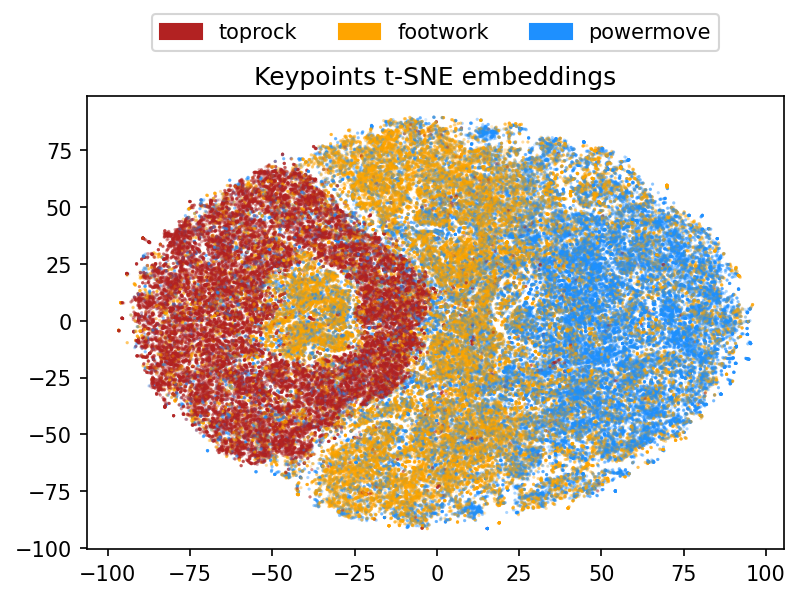}
    \caption{t-SNE embeddings of all keypoints in BRACE. A good separation of the three main dance elements emerges from static poses.}
    \label{fig:pose_tsne}
\end{figure}

We follow AIST++~\cite{li2021learn} and report the commonly used Fréchet Inception Distance~\cite{heusel2017gans}~(FID) calculated directly in the nodes pose  
space. More precisely, we first normalise keypoints as explained in Section~\ref{sec:dataset_comparison}. To calculate the FID, we stack all keypoints 
and estimate their mean and covariance.   
We report two dance-music correlation metrics introduced in~\cite{li2021learn}: Beat Alignment Score and Beat Dynamic-Time-Warping (DTW) Cost. These metrics measure the alignment and similarity between music and kinematics beats. Music beats were obtained using the Essentia library~\cite{essentia}. Kinematics beats were obtained by finding peaks in the averaged second-order derivatives of keypoint movements.

We also analyse the ability of the models to generate a sensible breakdancing sequence 
by looking at the distribution of the three dance elements. As we saw in Figure~\ref{fig:movements}, our sequences are mostly composed of \fw{} and \pwm{} movements, both around 37\% of all frames, with \tr{} movements amounting to around 23\% of the data. We check if the generated sequences reflect this distribution. 
We conduct this study by looking at frame poses. We first inspect whether static poses have a clear enough separation for our study to be sensible. We do this by plotting t-SNE embeddings~\cite{van2008visualizing} of all keypoints in BRACE. 
As depicted in Figure~\ref{fig:pose_tsne}, we observe a good separation of the three dance elements from static poses. This motivates us to train a classifier to recognise poses from single frames and use its predictions to estimate the distribution of dance elements in the generated sequences. Specifically, we train a spatial GCN model~\cite{2sagcn2019cvpr}, which 
attained 73.1\% top-1 accuracy on the test set. 

\subsection{Results}

\begin{table*}[t]
\centering
\caption{Testing recent generative  models on BRACE. Note that toprock, footwork and powermove averages for the GT reference baseline were predicted with the GCN classifier. Actual statistical average in the test set were (\tr{}=25.52, \fw{}=39.72, \pwm{}=34.76). $\downarrow$ lower is better, $\uparrow$ higher is better.}
\resizebox{\textwidth}{!}{%
\begin{tabular}{@{}lcccccc@{}}
\toprule
\textbf{Baseline} & \textbf{Pose FID} $\downarrow$ & \textbf{Beat alignment score} $\uparrow$ & \textbf{Beat DTW cost} $\downarrow$ & \textbf{Toprock avg} & \textbf{Footwork avg} & \textbf{Powermove avg} \\ \midrule
GT Reference & 0.0032 & 0.451 & 36.50 & 7.84 & 48.45 & 43.71 \\
Dance Revolution~\cite{huang2021} & 0.5158 & 0.264 & 11.88 & 10.59 & 51.60 & 37.72  \\ 
AIST++~\cite{li2021learn} & 0.5743 & 0.136 & 12.92 & 6.39 & 40.73 & 52.89  \\ 
Dancing 2 Music~\cite{lee2019dancing2music} & 0.5884 & 0.129 & 11.60 & 16.04 & 50.09  & 33.87 \\ \bottomrule
\end{tabular}%
}
\label{tab:results}
\end{table*}

We divide our dataset in a 70/30 training/test split and ensure all models reach training convergence before testing. Table~\ref{tab:results} reports results obtained on the selected models. 
The ``GT Reference'' baseline reports evaluation metrics calculated on the ground truth test set, i.e., on real dance sequences as opposed to generated ones. Since this is a new dataset, this helps us understand the results of the evaluated models. We observe that all baselines struggle to achieve optimal performance. In fact, pose FID and beat alignment score are far from the reference baseline, which indicates that the generated sequences are not very realistic. 
However, models are able to generate poses resembling breakdancing postures. This is indicated by the classifier predictions of \tr{}, \fw{} and \pwm{}. 
Nevertheless, except for AIST++, all models generate more \tr{} and fewer \pwm{} poses, which are respectively the easiest and hardest breakdancing postures. This demonstrates the challenges BRACE poses to existing state-of-the-art models when evaluated on complex scenes.
Dancing 2 Music achieves the worst FID and beat alignment score. As noted in Section~\ref{sec:rel_work_motion_syn}, this method splits sequences automatically into atomic dance units, and the model is trained to compose such units to form a dance sequence. Motion decomposition is done by measuring the displacement of each joint between neighbouring frames. Breakdancing moves with their large displacements  
stress this approach, which explains the poor performance of the method.

Interestingly, we notice that models attain a better beat DTW cost compared to the GT Reference baseline, but their beat alignment score is worse. Breakdancing songs feature fast-tempo music with regular beats. While dancer follow the tempo and flow of the music, naturally they do not perform a movement for every beat, hence the higher beat DTW cost of the reference baseline. This suggests models learn to ``blindly'' follow the music beat, i.e. they generate sequences full of strong movements which however do not necessarily mimic a real breakdancing, which typically displays a good mix of slow and fast moves.

Qualitative evaluation is as important as quantitative analysis for dance motion synthesis.
We visually inspect the generated sequences and notice that while motion appears plausible, ultimately sequences look disconnected, showing repeating movements. Although the pose classifier shows that models are able to generate toprock, footwork and powermove, we observe that the way these are mixed is not very realistic and diverse. This is most likely due to the weaker correlation of music and motion in breakdancing. 
The generated sequences also display motion induced by camera panning and zooming, which was expected since the tested methods assume the input motion data is controlled. 
We show these findings in the supplementary material video \footnote{\url{https://youtu.be/-5N6uBDfMoM}}. 

Lastly, we evaluate the impact of two hyper-parameters when training Dance Revolution~\cite{huang2021}, namely the number of layers (2-4) for the encoder and the sliding window size of the local self-attention (15, 25, 50, 100, 200). In Table~\ref{tab:results} we report results obtained with the best parameters, i.e. 3 layers and window size 15 (all results in supplementary material). The number of layers did not affect performance much. The attention size instead played an important role, with performance degrading as the window enlarged. This parameter controls the receptive field of the encoder, i.e. the temporal neighbourhood each element in the input sequence can attend to.  
Performance deterioration with larger windows is linked to the fact that our sequences contain motion with very quick and large displacement, which are hard to model well with a large receptive field.

\begin{table}[t]
\centering
\caption{Evaluating BRACE for pose estimation. We compare the performance of HRNet~\cite{sun2019deep} pre-trained on COCO~\cite{lin2014microsoft} before and after fine-tuning.}
\label{tab:pose_estimation_results}
\resizebox{0.9\textwidth}{!}{%
\begin{tabular}{@{}lcccccc@{}}
\toprule
\textbf{Model} & \textbf{BRACE AP} & \textbf{BRACE AR} & \textbf{Ext. powermoves AP} & \textbf{Ext. powermoves AR} \\ \midrule
COCO-pretrained & 0.158 & 0.202 & 0.462 & 0.513  \\
BRACE-finetuned & 0.357 & 0.445 & 0.598 & 0.642 \\ \bottomrule
\end{tabular}%
}
\end{table}

\section{Pose Estimation on BRACE}

We show here that BRACE 
can also be useful for 2D keypoint estimation since it contains well-labelled skeletons with motion blur and extreme poses.
We use the 26K manually annotated skeletons and split them into 80/20 train/test
sets. We use HRNet~\cite{sun2019deep} pretrained on COCO~\cite{lin2014microsoft} as base
model and compare performance before and after finetuning on BRACE. Table~\ref{tab:pose_estimation_results} reports a large improvement after fine-tuning. While this was expected, it proves manual annotations are consistent and models can learn from the data.
The poor performance of COCO-pretrained also corroborates that our pipeline successfully detected bad automatic key-points: frames are then manually labelled but have complex postures, so performance before finetuning is low. 
Besides showing improvements on the test split of BRACE, the BRACE-finetuned model also outperforms COCO-pretrained when tested on an external powermove video \footnote{\url{https://www.youtube.com/watch?v=q5Xr0F4a0iU}}. This video is a compilation of powermoves shot with a very different setting compared to BRACE. We randomly sample 200 frames from the video and manually annotate these frames. Although camera movements and angles as well as lighting in the external video are very different from those in our dataset, the BRACE-finetuned model shows a substantial improvement. This proves that BRACE is useful to help the model learn pose estimation in extreme postures.

\section{Conclusion}
\label{sec:conclusion}

We presented BRACE, a new dataset for audio-conditioned dance motion synthesis. BRACE was collected to challenge the main assumptions taken by motion synthesis models, i.e. relative simple poses and movement captured with controlled data. 
We designed an efficient pipeline to annotate poses in videos, striking a good balance between labelling cost and keypoint quality. Our pipeline is flexible and can be adopted for any pose estimation task involving complex movements and dynamic recording settings. We further enrich our dataset with temporal segments labelling the constituent elements of breakdancing. 
BRACE, while restricted to a specific dance genre, readily pushes models to go beyond a modality mapping approach, to reason more efficiently about motion and poses and to deal with less constrained scenarios.

\fparagraph{Future research} 
Because audio cues are not as strong as in other datasets, future models will have to creatively find other ways to generate good results on BRACE. For example, the movement labels could be exploited.  
This could be achieved by injecting the movement category in a model through positional encoding, i.e. by modelling the relative position of a movement in the sequence together with its type. Knowing that a given dance segment is a specific movement, models could generate more varied sequences by enforcing a given distribution or order of the movement categories. This is just one of the possible ideas one could develop with our fine-grained labels. 
Another interesting direction for further research is the generation of dance sequences in a finer granularity and more controlled manner. Current methods lack the ability to adjust generated sequences according to human intervention or conditioning. If we could decompose sequences into shorter actions, e.g. by effectively clustering poses and movements, 
we could then introduce user input as another modality to generate dance movements that follow a designated pattern. 
For example, a user might specify a target combination of toprocks, footworks and powermoves. 
Such work would be extremely helpful for animation and gaming industries. Since breakdancing is one of the most granular dance genre, we believe BRACE constitutes a good dataset to experiment such methods on.

\fparagraph{Longevity of BRACE}
While it is common practice to utilise YouTube videos to compile research datasets (e.g. Kinetics~\cite{carreira2017quo}, AVA~\cite{gu2018ava}, YouTube8M~\cite{abu2016youtube}), this comes with a risk. Because videos can disappear from YouTube and since researchers in most cases cannot publish their own copy of the videos, YouTube-based datasets are sometimes difficult to compare results on, and missing videos intrinsically diminish the usefulness of a dataset. We are aware of this issue, however for dance motion synthesis, the main scope of our work, this problem is not as severe. We release our keypoint sequences together with Librosa~\cite{mcfee2015librosa} audio features, thus even if all the corresponding YouTube videos were removed, our dataset would still be entirely usable and future generative models can be compared accordingly. We note this is common practice in other skeleton-based datasets for dance synthesis~\cite{lee2019dancing2music,huang2021,ferreira2021learning} and we follow this approach accordingly.

\noindent \textbf{Acknowledgement}. This study is supported under the RIE2020 Industry Alignment Fund - Industry Collaboration Projects (IAF-ICP) Funding Initiative, as well as cash and in-kind contribution from the industry partner(s). The project is also supported by Singapore MOE AcRF Tier 1 (RG16/21).

\pagebreak

\appendix

{\noindent\Large\textbf{Supplementary Material}}

\section{Video} 

We prepared a video to showcase our dataset. For a better viewing we outline the video below. We overlay our keypoints onto the original videos, however we only release keypoints since video copyright belongs to Red Bull, to which we are not affiliated. The video is available at \url{https://youtu.be/-5N6uBDfMoM}.

\paragraph{Active dancer detection} We show the input/output of our active dancer detection algorithm, i.e. all the poses obtained with the pose estimation models and the automatically detected performing dancer. Notice the abundant number of boxes and keypoints and the accurate detection of the active dancer.

\paragraph{Merging and interpolating sequences} We show a sequence of raw automatic poses and the corresponding sequence obtained merging and interpolating automatic and manual keypoints. Notice here the noisy estimations in the raw sequence. Noisy poses were automatically rejected, and a few of them were manually labelled. The final sequence merges good automatic poses and manually annotated keypoints, interpolating the rejected frames we did not annotate due to our discount method. This example shows how our approach is able to produce good sequences with a low annotation cost. We recommend slowing down the video during this section. We added a text colour legend to illustrate which keypoints were rejected or missing, manually annotated or were good automatic keypoints.

\paragraph{Dataset samples} We show a few toprock, footwork and powermove segments. We display the interpolated keypoint sequences and play the corresponding audio. Notice the main characteristics that make keypoint extraction challenging: multiple moving cameras, fast lighting changes, occlusions, tangled poses and fast movements. Complex postures and dynamic moves also make BRACE a unique data source for dance motion synthesis.

\paragraph{Experiment results} We compare sequences generated with the tested baselines: Dance Revolution~\cite{huang2021}, AIST++~\cite{li2021learn} and Dancing to Music~\cite{lee2019dancing2music}. The input for testing these models is an audio snippet. In the video we show the corresponding footage segment for reference. Here we provide a few comments about these examples: 

\begin{itemize}
    \item Dance Revolution generates sequences ``that never stop'', i.e. the dancer is constantly moving throughout the video. While this does not reflect a real breakdancing sequence very well, the continuous motion and the plausible movements help this model achieve the highest pose FID amongst the evaluated baselines.
    \item  AIST++ is able to produce sequences following a more natural tempo, i.e. it generates both fast and slow motions as well as a good mixture of toprock, footwork and powermove segments. However AIST++ also picks up much of the motion induced by the recording setting, i.e. moving/zooming/panning cameras and shot changes, which explains the lower pose FID obtained by the model. Note that due to the required inference seeding, the first 2 seconds of the sequences generated by AIST++ are identical to the ground truth (more details in Section~\ref{sec:setup}).
    \item Dancing to Music can generate reasonable movements, however these are severely disconnected and consequently the produced sequences do not look realistic. This is due to the automatic decomposition of the dance as we explained in the main text, and is the main cause of the poorer performance of this baseline. This model too picks up camera-induced motion, although to a lesser extent compared to AIST++.
\end{itemize}

\noindent While in the supplementary video we show only a few examples, our findings apply to all the generated sequences we reviewed. 

\section{Experimental Setup}
\label{sec:setup}

When testing models on BRACE we ensured training convergence and sensible test output. Here we provide details about our experimental setup. We used the same acoustic features extracted with Librosa~\cite{librosa} following~\cite{huang2021} for all baselines.

\paragraph{Dance Revolution~\cite{huang2021}} For training we set learning rate to $1\text{e-}4$ and batch size to 32. We evaluated different numbers of layers in the encoder, namely 2, 3 and 4, as well as different sizes for the self-attention window: 15, 25, 100, 200. Training sequence length was set to 200 frames (longer sequences are split accordingly). Pose and frame embedding sizes were set to 34 (number of nodes times 2) and 200 respectively. The number of hidden units for the decoder was set to 1024. The condition step $q$ and $\lambda$ parameters for the dynamic auto-condition learning scheme were set to 10 and 0.1. Please refer to~\cite{huang2021} for details about these hyper-parameters.

\paragraph{AIST++~\cite{li2021learn}} For training we follow the implementation details reported in~\cite{li2021learn}, except for the sequence length. In~\cite{li2021learn} the authors used a seed motion sequence of 120 frames and a music sequence of 240 frames to predict the future 20 frames, where L2 loss is used to optimise the network. However, since their videos are 60 fps while ours are mostly 25 fps, we adjust the seed sequence length to be 50 frames for motion and 100 frames for music, and supervise the L2 loss with the future 8 frames. 
AIST++ requires a seeding sequence for inference as well.
For testing we use the first 50 frames of the motion sequence from the ground truth to generate future frames,  hence in our supplementary video it can be observed that the first two seconds of AIST++'s sequences are identical to the ground truth. We exclude these initial 50 frames when calculating the reported metrics. 

\paragraph{Dancing to Music~\cite{lee2019dancing2music}} 
We split sequences into segments of 32 frames (roughly 1.2 seconds at 25 fps) using kinematic beats. These were obtained by finding peaks in second order derivatives of keypoint displacements. We used the same approach to find kinematic beats to calculate the beat alignment score and DTW cost, as mentioned in the paper. We use the split segments to train the decomposition network. Specifically, we extract the first four segments from each sequence. These four segments together with the corresponding music features form a training sample for the composition network. The used hyper-parameters are the same as those specified in the official code repository.

\begin{table*}[t]
\caption{Extended results obtained with Dance Revolution~\cite{huang2021}. Here we show the impact of varying the number of layers in the encoder as well as the size of the self-attention window.}
\centering
\resizebox{\textwidth}{!}{%
\begin{tabular}{@{}cccccccc@{}}
\toprule
\textbf{Layers} & \textbf{Attention size} & \textbf{Pose FID} $\downarrow$ & \textbf{Beat alignment score} $\uparrow$ & \textbf{Beat DTW cost} $\downarrow$ & \textbf{Toprock avg} & \textbf{Footwork avg} & \textbf{Powermove avg} \\ \midrule
\multirow{5}{*}{2} & 15 & 0.5239 & 0.269 & 11.89  & 6.92 & 54.20 & 38.88 \\
 & 25 & 1.1174 & 0.273 & 11.80 & 6.12 & 55.82 & 38.06 \\
 & 50 & 0.9663 & 0.271 & 11.75 & 6.52 & 55.53 & 37.95 \\
 & 100 & 1.0816 & 0.266  & 11.70  & 6.70 & 55.59 & 37.71 \\
 & 200 & 0.7924 & 0.265 & 11.67 & 11.11 & 52.13 & 36.77 \\ \midrule
3 & 15 & 0.5158 & 0.264 & 11.88 & 10.59 & 51.69 & 37.72 \\ \midrule
4 & 15 & 0.5723 & 0.263 & 11.85 & 10.25 & 52.64 & 37.12 \\ \bottomrule
\end{tabular}%
}
\label{tab:dance_rev_all_results}
\end{table*}

\section{Dance Revolution extended results}

Table~\ref{tab:dance_rev_all_results} reports extended results obtained with Dance Revolution~\cite{huang2021}. We trained the model varying the number of layers in the encoder as well as the size of the self-attention window. While the number of layers plays a limited role (best results obtained with 3 layers), the attention window size has a stronger impact on performance (best results with size 15). In particular, we observe that pose FID degrades as the window is enlarged. This parameter controls the receptive field of the encoder, i.e. the temporal neighbourhood each element in the input sequence can attend to. 
Since our sequences display complex motion with large displacement, a smaller receptive field is best suited at modelling the dance patterns. 
We do not see noticeable changes in beat alignment score and beat DTW cost, which suggests that the ability of the network to align dance and music is not affected by the number of encoding layers and the attention window size.

\begin{figure}[t]
    \centering
    \includegraphics[width=\linewidth]{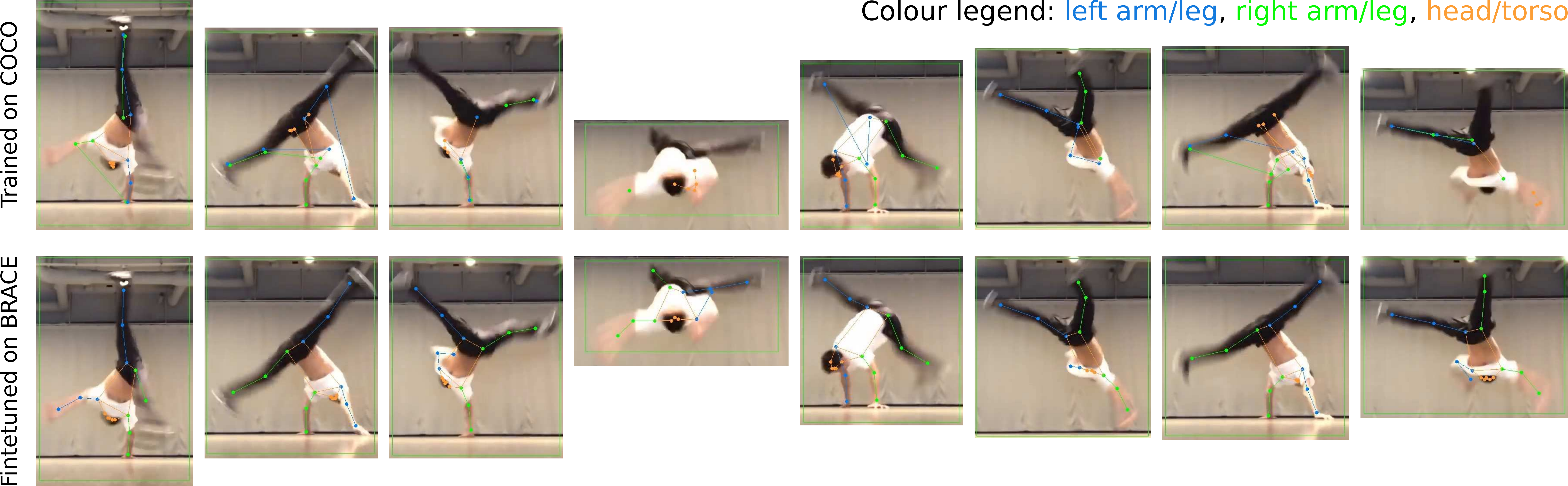}
    \caption{Qualitative results for pose estimation on the external powermoves video.}
    \label{fig:pwmv_qual}
\end{figure}

\section{Pose Estimation}

We show here qualitative results for the pose estimation experiment on the external powermoves video \footnote{\url{https://www.youtube.com/watch?v=q5Xr0F4a0iU}}. Figure~\ref{fig:pwmv_qual} compares poses estimated with the base model (HRNet~\cite{sun2019deep}) trained on COCO~\cite{lin2014microsoft} (top row) and the model finetuned on BRACE (bottom row). 
We display here particularly difficult cases where the base model struggled due to the complex pose and the prominent motion blur. Notice how the model finetuned on BRACE is able to predict very accurate keypoints thanks to the additional training on our manually labelled frames. Our data processing pipeline picks the hardest frames for manual labelling. Our high quality labels for such difficult frames prove effective for the model to learn to predict accurate keypoints even for very tangled postures in the presence of motion blur. While the main scope of our work is dance motion synthesis, we believe BRACE can also be a valuable resource for pose estimation.

\section{Bézier interpolation}

A \bez{} curve is a parametric curve based on Bernstein polynomials. Given a set of control points ${P = (P_0, P_1, \dots, P_{d}})$,
a \bez{} curve can be obtained multiplying the Bernstein matrix by the control points. For a given order $d$ and a vector $\tau$ of $m$ values uniformly spaced between $0$ and $1$, the Bernstein matrix $M \in \mathrm{R}^{m \times (d+1)}$ is defined as follows:

\begin{equation}
M_{i, j+1} = \binom{d}{j} (1-\tau_i)^{d-j} \tau_i^j 
\label{eq:bernstein_matrix}
\end{equation}

\noindent where $0 \le i < m$, $0 \le j \le d$ and $\tau \in [0, 1]$. 
In our case, $P$ is unknown and is what we need to find to fit a \bez{} curve to a node trajectory. 
Let $R \in \mathrm{R}^{q \times 2}$ be the matrix formed stacking the 2D coordinates of a node trajectory, i.e. ${R_i = (x_i, y_i)}$.
$R$ can contain a mix of automatic and manually annotated 2D points. Note that keypoints are not expected to be temporally continuous due to missing or discarded poses. In other words, if $T(i)$ is a function returning the corresponding frame number of the $i$-th point in the trajectory, $T(i+1) = T(i) + 1$ does not necessarily hold true. 
Following~\cite{pastva1998bezier} we compute $P$ using the least square method. Namely, we first calculate $M$ with a $\tau$ vector of $q$ values and predefined order $d$. We then calculate the Moore–Penrose inverse $M^+$ of $M$. Finally, the fitted control points are given by $P = M^+ R$.
Once $P$ is obtained we can generate the interpolated trajectory by multiplying the previously calculated $M$ by $P$. 
We adopt a sliding window approach to interpolate keypoint sequences. Specifically, we slide a window of 15 frames with stride 14. The overlapping frame between windows allows for a smoother transition. We set the degree of the curve to 7.

%
%
\bibliographystyle{splncs04}

\end{document}